# Evidence-based Anomaly Detection in Clinical Domains

**Milos Hauskrecht, PhD[1-3], Michal Valko, MSc[1], Branislav Kveton, PhD[2,*], Shyam Visweswaran MD, PhD[2-3], Gregory F. Cooper, MD, PhD[2-3]**
**[1] Computer Science Department; [2] Intelligent Systems Program; [3] Department of Biomedical Informatics, University of Pittsburgh, PA**

**Abstract**

*Anomaly detection methods can be very useful in identifying interesting or concerning events. In this work, we develop and examine new probabilistic anomaly detection methods that let us evaluate management decisions for a specific patient and identify those decisions that are highly unusual with respect to patients with the same or similar condition. The statistics used in this detection are derived from probabilistic models such as Bayesian networks that are learned from a database of past patient cases. We evaluate our methods on the problem of detection of unusual hospitalization patterns for patients with community acquired pneumonia. The results show very encouraging detection performance with 0.5 precision at 0.53 recall and give us hope that these techniques may provide the basis of intelligent monitoring systems that alert clinicians to the occurrence of unusual events or decisions.*

**Introduction**

Patient medical records today include thousands of electronic entries related to patient conditions and treatments. While these have proven useful in providing a better picture about the individual patient, the benefits of such data in decision support, or in discovery and acquisition of new clinical knowledge are far from being exhausted. The objective of this research is to develop computational tools that utilize previously collected patient data to detect unusual patient-management patterns. The hope is that these methods can eventually lead to systems that will alert clinicians to unusual treatment choices. Such systems would have the advantage of not requiring labor-intensive extraction and encoding of expert knowledge. Additionally, we envision these methods will enable knowledge discovery, where unusual outcomes and their contexts are identified.

We propose and investigate a new statistical anomaly framework to detect unusual patient-management decisions based on a probabilistic model. Briefly, our approach builds a probabilistic model **M** that captures stochastic dependencies among attributes in the data. We use the Bayesian belief network framework [1,2,3,4] to compactly represent such dependencies. The probabilistic model is then used to compute a predictive statistic for a patient case that is based on the conditional probability $p(A \mid C, \mathbf{M})$ of the target features $A$ (representing patient-management decisions) given the values of other features $C$ (such as patient symptoms). The target patient (that is, the patient we want to evaluate) is then compared to similar patient cases and unusual or anomalous decisions are detected based on the differences in their predictive statistics.

Probabilistic anomaly detection methods are sensitive to the sample of patient cases used to build the model, both in the number of patient cases used, as well as, the extent of their similarity to the target patient case. Balancing both objectives in practice can be very hard. The number of cases in the data that match well the target case is often too small to provide sufficient support to draw any statistically sound inferences. On the other hand, the naive comparison with all patient cases may lead to detection errors. For example, a patient case may be declared as anomalous by a detection method simply because it falls into a low prior probability cohort. To correct for this problem we investigate *conditional anomaly detection methods* that search for a subpopulation of past patient cases that is adequate both in terms of the quality of the match to the target patient case as well as in its size, so that sound anomaly conclusions can be drawn. As an example application, we empirically evaluate the performance of our methods on the problem of detection of unusual hospitalization patterns for patients with community acquired pneumonia.

**Methodology**

In anomaly detection, we are interested in detecting an event (a case) the occurrence of which deviates from past cases (or events). Let $\mathbf{x}$ be a patient case that we want to analyze and determine whether it is anomalous in terms of some target attribute (or a set of target attributes) $A(\mathbf{x})$. Let $\mathbf{E}=\{\mathbf{x}^1, \mathbf{x}^2, \ldots \mathbf{x}^n\}$ be a

* Branislav Kveton contributed to this work while he was a graduate student at the University of Pittsburgh. He is currently with Intel Research.

set of past cases similar to **x** and **M** be a probabilistic model that represents the distribution of cases in **E**.

To assess the anomaly we analyze a predictive statistic of A(**x**) obtained for the target case **x** with respect to **M** and the remaining (non-target) attributes C(**x**). We say the case **x** *is anomalous* in the target attribute(s) A(**x**), if the probability $p(\mathrm{A}(\mathbf{x}) \mid \mathrm{C}(\mathbf{x}), \mathbf{M})$ is small and falls below some threshold.

To build a working anomaly detection algorithm, we need to provide answers to the following three questions: (1) How should the probabilistic model **M** be constructed? (2) How should the anomaly be detected using the predictive statistic? (3) How should cases **E** that are similar to the case **x** be selected? We provide answers to these questions in the remainder of the paper.

**Building a probabilistic model**: The calculation of $p(\mathrm{A}(\mathbf{x}) \mid \mathrm{C}(\mathbf{x}), \mathbf{M})$ assumes the existence of an underlying probabilistic model **M** that describes stochastic relations among the attributes. However, the number of attributes in real-world datasets, especially those that are collected in medical applications, can be enormous. In general, it is not feasible to represent the model by enumerating the full joint probability distribution since its complexity grows exponentially in the number of attributes. To address this concern, we adopt succinct probabilistic representations, in particular, the *Bayesian belief network (BBN)* model [1,2], and its special instance: the Naïve Bayes classifier model [5].

A BBN model **M** is represented by a pair $(S_M, \theta_M)$, where $S_M$ denotes the model structure and $\theta_M$ its parameters. We adopt the Bayesian framework [2] to learn the parameters of the model and to compute any related statistics. In this framework the parameters $\theta_M$ of the model are treated as random variables and are described in terms of a density function $p(\theta_M \mid S_M)$. The probability of an event $u$ is obtained by averaging over all possible parameter settings of the model **M**:

$$p(u \mid S_M) = \int_{\theta_M} p(u \mid S_M, \theta_M) p(\theta_M \mid S_M) d\theta_M$$

To incorporate the effect of cases **E**, $p(\theta_M \mid S_M)$ corresponds to the posterior $p(\theta_M \mid \mathbf{E}, S_M)$, which is obtained via Bayes rule:

$$p(\theta_M \mid \mathbf{E}, S_M) = p(\mathbf{E} \mid \theta_M, S_M)\, p(\theta_M) \,/\, p(\mathbf{E}, S_M),$$

where $p(\theta_M)$ defines the prior for parameters $\theta_M$. To simplify calculations, we assume (1) parameter independence and (2) conjugate priors [2]. With these assumptions, the posterior follows the same distribution as the prior and the computation reduces to updates of sufficient statistics. Similarly, many of the probabilistic calculations can be performed in closed form.

**Anomaly detection**: Multiple threshold approaches can be used to make anomaly calls based on the predictive statistic. These include: absolute thresholds, relative thresholds, and the *k*-standard-deviation thresholds. In our work, we build upon the *absolute threshold test*. In the *absolute threshold test*, the case **x** is anomalous if $p(\mathrm{A}(\mathbf{x}) \mid \mathrm{C}(\mathbf{x}), \mathbf{M})$ falls below some fixed probability threshold $p_\varepsilon$. Intuitively, if the probability of the target attributes A(**x**) for **x** is low with respect to the model **M** and its other attributes C(**x**), then the value of the target attribute is anomalous. Note that the absolute threshold test relies only on the model **M** and there is no direct comparison of the predictive statistic for **x** with those of the cases in **E** that are similar to it. However, recall that the cases in **E** are used to construct the model **M** and hence their effect is reflected in the statistic.

**Selection of similar cases**: Probabilistic detection criteria are sensitive to the choice of cases **E** used to learn the model **M**. A simple solution is to use all cases in the data repository in constructing the set **E**. The problem with this approach is that the resulting statistic is likely to be biased towards cases that are more frequent in the dataset, which in turn can affect the statistic used in the detection criterion. For example, if the target case happens to fall into a small subpopulation of patient cases, the management choice that is routinely made for the patients in this group, but rarely outside of this group, can be declared anomalous simply because the subpopulation is relatively small.

A solution to the above problem of false positives is to identify a subpopulation of cases that is relevant to the target case. Only these cases are then used in anomaly detection. However, this rather intuitive approach leaves two open questions: How should cases most relevant to the target case be selected? What is the minimal size of **E** sufficient to warrant any statistically sound inferences?

Clearly, the best subpopulation is the one that exactly matches the attributes C(**x**), of the target case, that is, $\mathbf{E} = \{\mathbf{x}^i\text{: such that } \mathrm{C}(\mathbf{x}^i) = \mathrm{C}(\mathbf{x})\}$. However, in real-world databases few if any past cases are likely to match the target case exactly; so there is little basis on which to draw statistically sound anomaly inferences. To meet the minimum size condition, we propose strategies based on similarity metrics that relax the exact match criterion.

**Similarity-based match:** The distance (similarity) metric defines the proximity of any two cases in the dataset. We define a similarity metric on the space of attributes C(**x**) that lets us select cases closest to the target case **x**. The *k* closest matches to the target case then define the best subpopulation of size *k*. Different distance metrics are possible. An example is the generalized distance metric $r^2$ defined as: $r^2(\mathbf{x_i},\mathbf{x_j}) = (\mathbf{x}_i - \mathbf{x}_j)\Gamma^{-1}(\mathbf{x}_i - \mathbf{x}_j)^T$, where $\Gamma^{-1}$ is a matrix that weights attributes of patient cases in proportion to their importance. Different weights lead to a different distance metrics. For example, if $\Gamma^{-1}$ is the identity matrix **I**, the equation defines the Euclidean distance of $\mathbf{x}_i$ relative to $\mathbf{x}_j$. The Mahalanobis distance [6] is obtained by setting $\Gamma$ to the population covariance matrix $\Sigma$ which lets us incorporate dependencies among the attributes.

**Attribute importance:** In practice, some of the attributes are more important for defining the similarities than others. For example, attributes that are known to be correlated with the target attribute should be accorded more importance than attributes that are not. To address this issue we propose to re-weight the Mahalanobis distance metric according to the ability of each attribute to predict the target attribute, so that attributes influencing the target attribute matter more.

We define the weighted Mahalanobis distance as:

$$r_w^2(\mathbf{x_i},\mathbf{x_j}) = \mathbf{w} * (\mathbf{x}_i - \mathbf{x}_j)\Gamma^{-1}(\mathbf{w} * (\mathbf{x}_i - \mathbf{x}_j))^T,$$

where * defines the element-wise multiplications of the two vectors. The importance weight vector **w** can be obtained using a variety of scoring metrics that measure the strength of (univariate) dependencies between the attributes *C* and the target attribute *A*.

The similarity-based population selection method finds a subpopulation of cases in the data that consists of *k*-best matches to the target case. However, it is still unclear if the matches are of good quality. To address this we calculate, for each case in the subpopulation, its average distance from its *k*-best matches and reject the subpopulation for detection purposes if the average distance for the target case deviates significantly from the average distances observed for all other cases. We use a threshold of 2 standard deviations for making this call.

**Experimental evaluation**

To evaluate the potential of the our anomaly detection framework, we applied it to the Patient Outcomes Research Team (PORT) cohort study's dataset [7,8] to detect unusual patient admission decisions. The PORT dataset has data on 2287 patients with community acquired pneumonia that were collected in a study conducted from October 1991 to March 1994 at five medical institutions. The original PORT data were analyzed by Fine et al. [8], who derived a prediction rule for predicting 30-day mortality. To explore data-driven detection methods, we experimented with a simpler version of the PORT dataset that contains, for each patient, only those attributes identified as most relevant by Fine's study. These attributes are summarized in Table 1. All attributes are binary with true / false (positive / negative) values. Our objective was to detect *unusual admission decisions* (treat the patient at home versus in the hospital) which are captured by the variable "Hospitalization".

| | |
|---|---|
| **Target attribute:** | **Physical-examination:** |
| Hospitalization | Pulse ≥125 per min |
| | Respiratory rate ≥ 30 per min |
| **Demographic factors** | Sys. blood pressure < 90mm Hg |
| Age > 50 | Temperature < 35C or > 40C |
| Gender (male, female) | |
| | **Lab. & radiograph. findings:** |
| **Co-existing illnesses:** | Blood urea nitrogen $\geq 30$ mg/dl |
| Congestive heart failure | Glucose $\geq 250$ mg / dl |
| Cerebrovascular disease | Hematocrit < 30 % |
| Neoplastic disease | Sodium < 130 mmol / l |
| Renal disease | Art. $O_2$. pressure < 60 mm Hg |
| Liver disease | Arterial pH < 7.35 |
| | Pleural effusion |

**Table 1.** Attributes of the PORT dataset.

**Study design:** To evaluate the performance of our anomaly detection methods, we used 100 patient cases (out of a total of 2287 of cases). The chosen cases included 21 cases that were found to be anomalous according to the Naïve Bayes detector (using a 5% detection threshold) that was trained on all data but the target case (i.e. 2286 cases); the remaining 79 cases were selected randomly from the rest of the dataset. Each of the 100 cases was then evaluated independently by each member of a panel of three physicians. The physicians were asked whether they agreed with the hospitalization decision for a patient case based on the values of the attributes listed in Table 1. Using the panel's answers, the admission decision was labeled as anomalous when (1) at least two physicians disagreed with the actual admission decision that was taken for a given patient case or (2) all three indicated they were unsure about the appropriateness of the management decision. We used these labels as the gold standard for defining anomalous hospitalization decisions. Out of the 100 cases, the panel judged 23 as anomalous hospitalization decisions, and the remaining 77 as not anomalous.

**Experiments:** All anomaly detection experiments that we performed followed the *leave-one-out* scheme. That is, for each case in the dataset of 100 patient cases evaluated by the panel, we identified a set of cases **E** most similar to it in the PORT dataset while excluding the just evaluated case, and used them to train one of the probabilistic models (see below). Given the trained model, the posterior probability of the target case was calculated. The target case was declared anomalous if its posterior probability value fell below the detection threshold. The anomaly determinations made by our algorithms were compared to the assessment of the panel and evaluation statistics (sensitivity, specificity, precision) were calculated. To gain insight into the overall performance of each method, we varied its detection threshold and calculated the corresponding Receiver operating characteristic (ROC) and Precision-Recall (PR) curves and the areas under these curves.

**Probabilistic models**: We used two different BBN models: (1) the *Naïve Bayes model* and (2) a BBN model whose structure was learned from the PORT dataset. The Naive Bayes model [5] was constructed such that the target decision variable "Hospitalization" was represented by the class variable. The parameters of the model were learned from patient cases chosen by the population-selection method. To learn the structure of the second model, we used all but the 100 patient cases (= 2187 cases) selected for the evaluation and identified the structure with the maximal marginal likelihood score [4] using greedy BBN construction methods [3]. The structure was then kept fixed and the parameters of the model were learned from the cases picked by the population-selection method.

**Population selection:** We applied three different population-selection methods to learn the parameters of the model **M**: (1) the unrestricted population (excluding only the target case), (2) the 40 best matches based on the standard Mahalanobis distance, and (3) the 40 best matches based on the weighted Mahalanobis metric. The weight of each attribute was based on the Wilcoxon ranksum statistic that measures the ability of that attribute to predict the target "Hospitalization" variable.

## Results

Table 2 summarizes the performance results obtained for different anomaly detection methods The summary statistics listed are the area under the ROC and the area under the PR curve. The ROC analysis lets us analyze the performance of the detection model under different misclassification error tradeoffs. The PR curve and its area statistic are typically used in the evaluation of information retrieval systems and reflect the tradeoff between sensitivity (recall) and the true alarm rate (positive predictive value). Figure 1 illustrates the PR curve for one of the models: the BBN-based model trained on cases with the weighted Mahalanobis fit. Figure 2 shows the sensitivity and specificity for different detection thresholds for the same model.

| Prob. model | Population selection method | AUC PR | AUC ROC |
|---|---|---|---|
| NB | All cases | 0.42 | 0.74 |
| NB | Mahalanobis | 0.49 | 0.77 |
| NB | Weighted Mahalanobis | 0.60 | 0.80 |
| BBN | All cases | 0.56 | 0.81 |
| BBN | Mahalanobis | 0.52 | 0.79 |
| BBN | Weighted Mahalanobis | 0.56 | 0.80 |

**Table 2.** Comparison of probabilistic anomaly detection models based on the Naïve Bayes and the learned BBN for various population-selection criteria.

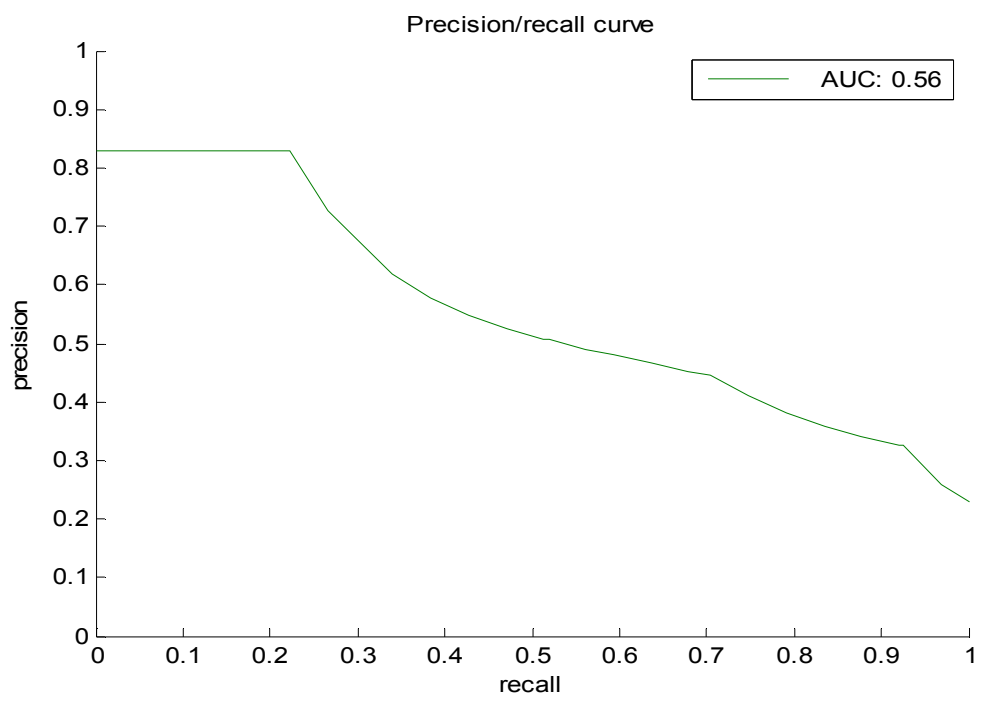

**Figure 1.** The Precision-Recall (PR) curve for the BBN model with the weighted Mahalanobis population selection.

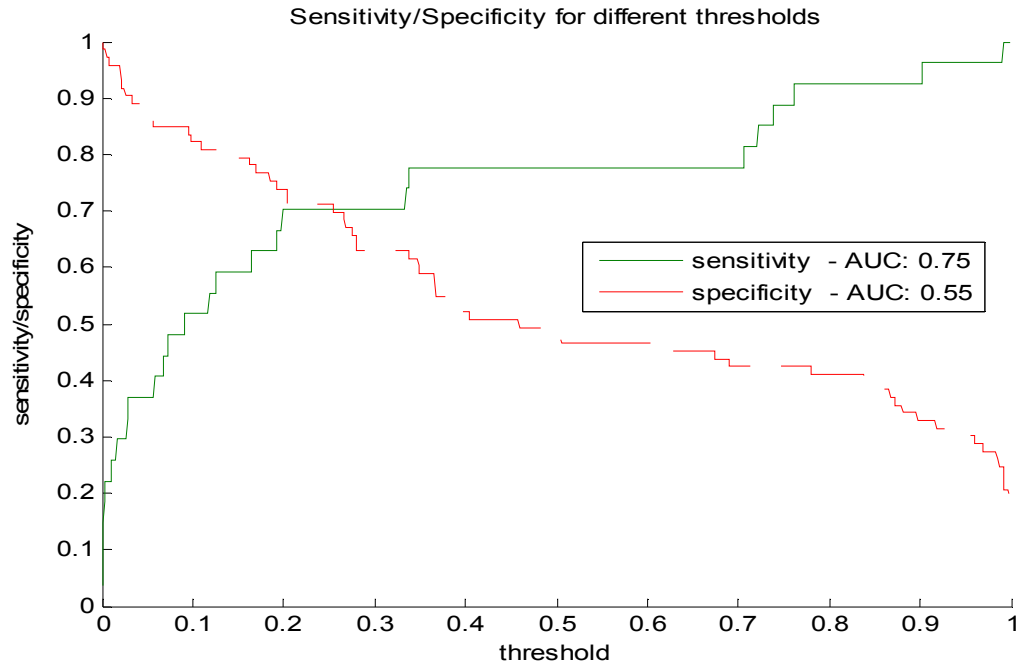

**Figure 2.** Sensitivity and specificity for different anomaly detection thresholds.

## Discussion

The Naïve Bayes model trained on all patient cases represents a baseline detection model. The model was easily outperformed by the BBN model also trained on all cases. This result can be explained by the ability of the BBN structure learning procedure to capture additional conditional independence relations among attributes which in turn helps to better assess the probability of the decision.

The similarity-based subpopulation selection leads to improved detection performance when it is combined with the Naïve Bayes model. In terms of the similarity metrics tested, the weighted Mahalanobis metric outperforms the standard Mahalanobis metric. This implies that the attributes that are not important for predicting the target decision should be weighted less than those with high predictive ability.

The BBN-based detection model does not appear to benefit from the population-selection enhancement when applied to PORT data. For the standard Mahalanobis metric, we even observe a drop in the model's detection performance. These results are likely due to higher complexity of the BBN model and a relatively small sample (40 best matches) that is used to train it. While a smaller sample may provide a better fit to the target case, it also decreases the accuracy of the parameter estimates of the model, especially if the model is more complex. This demonstrates the key trade-off of our population-selection enhancement. In terms of the two similarity metrics, we can once again observe the benefit of the metric re-weighted by the attributes' predictive performance, which, similar to the Naïve Bayes model, leads to better detection performance.

The PR curves and statistics indicate the precision of our methods to be 0.5 (1 correct anomaly in 2 anomalies reported) at 0.53 recall, which is very encouraging. However, given the method we used to construct the 100 test cases, the proportion of anomalies in the test data is likely inflated. Hence, the precision related statistics calculated from the data may be biased. Note that this, however, is not a problem for the relative comparison of the methods in Table 1, but can lead to an overestimate of the precision. Applying a conservative estimate of the prior occurrence of anomalies in the PORT dataset that is based on the incidence of anomalies in the randomly selected portion of the data, we estimate the precision of the method on the PORT data to be 0.32 (1 correct anomaly in every 3 anomalies reported) at 0.3 recall, which is still very encouraging.

## Conclusions

Statistical anomaly detection is a promising methodology for detecting unusual events that may correspond to medical error or unusual clinical outcomes. The advantage of the method over standard error detection approaches based on expert-extracted rules is that it works fully unsupervised and with little input from domain experts. Since it is complementary to rule-based anomaly detection, they could be applied jointly. A limitation of the method is that it is based on the conjecture that actions that are anomalous based on prior local practice are worth raising as alerts. The potential of the method is demonstrated on the PORT dataset with respect to anomalous hospitalization decisions. The initial results are promising and several further refinements of the approach remain to be investigated. For example, our current research aims to eliminate the effect of small subpopulation size on the parameter estimation process through subpopulation smoothing methods.

## Acknowledgements

This research was funded by grants R21-LM009102 and R01-LM008374 from the National Library of Medicine, and grant IIS-0325581 from the National Science Foundation.